\documentclass[runningheads]{llncs}
\usepackage[T1]{fontenc}
\usepackage{graphicx}
\usepackage{chngcntr}
\usepackage{siunitx}
\usepackage{tikz}
\usepackage{adjustbox}
\usepackage{url}
\usepackage{subfigure}
\usepackage{balance}
\usepackage{hyperref}
\usepackage{mathtools}
\usepackage[noend,linesnumbered]{algorithm2e}
\usepackage{algpseudocode}
\usepackage{multirow}
\usepackage{amssymb}
\usepackage{pifont}
\usepackage{booktabs}
\usepackage{array}
\usepackage{colortbl}
\usepackage{cite}
\usepackage{cleveref}

\Crefname{figure}{Fig.}{Figs.}
\Crefname{equation}{Eq.}{Eqs.}
\Crefname{section}{Sec.}{Secs.}

\algdef{SE}[DOWHILE]{Do}{doWhile}{\algorithmicdo}[1]{\algorithmicwhile\ #1}%

\RestyleAlgo{ruled}

\newcolumntype{L}[1]{>{\raggedright\let\newline\\\arraybackslash\hspace{0pt}}m{#1}}
\newcolumntype{C}[1]{>{\centering\let\newline\\\arraybackslash\hspace{0pt}}m{#1}}
\newcolumntype{R}[1]{>{\raggedleft\let\newline\\\arraybackslash\hspace{0pt}}m{#1}}

\newcommand{\cmark}{\ding{51}}%
\newcommand{\xmark}{\ding{55}}%

\begin{document}
\title{On the Fly Adaptation of Behavior Tree-Based Policies through Reinforcement Learning}
\titlerunning{On the Fly Adaptation of Behavior Tree-Based Policies}
%
\author{Marco Iannotta\orcidID{0000-0002-2142-6516} \and
Johannes A. Stork\orcidID{0000-0003-3958-6179} \and
Erik Schaffernicht\orcidID{0000-0002-0804-8637} \and
Todor Stoyanov\orcidID{0000-0002-6013-4874}}
\authorrunning{M. Iannotta et al.}
%
\institute{Center for Applied Autonomous Sensor Systems (AASS),\\ Örebro University, Sweden
\email{[firstname.lastname]@oru.se}}
\maketitle              
\begin{abstract}
With the rising demand for flexible manufacturing, robots are increasingly expected to operate in dynamic environments where local disturbances—such as slight offsets or size differences in workpieces—are common.
We propose to address the problem of adapting robot behaviors to these task variations with a sample-efficient hierarchical reinforcement learning approach adapting Behavior Tree (BT)-based policies.
We maintain the core BT properties as an interpretable, modular framework for structuring reactive behaviors, but extend their use beyond static tasks by inherently accommodating local task variations.
To show the efficiency and effectiveness of our approach, we conduct experiments both in simulation and on a Franka Emika Panda 7-DoF, with the manipulator adapting to different obstacle avoidance and pivoting tasks.
\end{abstract}
\section{Introduction}
In recent years, there has been a growing demand for intelligent autonomous systems to drive the adoption of flexible manufacturing processes in industry.
This shift requires that factory robots are able to adapt to the inherent variability of dynamic real-world scenarios and are imbued with intelligence beyond simple reproduction of preprogrammed behaviors.
While some situations invariably would require that an intelligent robot change the high-level decision logic it follows, many scenarios only feature local disturbances --- e.g., a slight offset or size difference of a workpiece the robot is picking.
In this paper, we consider such local disturbances, or \textit{task variations}, and propose a method to adapt the robot's control policy in response.

Behavior Trees (BTs) have emerged as a powerful tool in the field of Artificial Intelligence for representing and controlling the behavior of autonomous agents and robots~\cite{bts_book,bts_survey}.
BTs offer a hierarchical and modular framework that facilitates the design of reactive behaviors in dynamic environments.
Recent work has shown the feasibility of learning BT-based policies that work well under controlled and static conditions~\cite{learn_params_bts,styrud2024bebop}.
Compared to black-box policies (e.g. neural networks), BT-based policies not only provide interpretability of the robot's decision-making process but are also known to be faster and safer to learn~\cite{learn_params_bts,styrud2024bebop}.

However, adapting BT-based policies to novel task variations is challenging.
Unlike black-box approaches often trained to be robust under different conditions, BT-based policies are usually designed or learned with a single task in mind.
To our knowledge, Ahmad et al.~\cite{perf} is the only work addressing the problem of adapting BT-based policies to novel task variations.
Yet, their approach requires solving an optimization problem from scratch for a representative number of task variations, which can be prohibitive for problems with high-dimensional action spaces or a large space of variations.
As the task variations we consider are generally \textit{local} phenomena, there is a strong case for exploring methods that transfer experience between task variants during learning.

Our main contribution is an efficient method for learning how to adapt BT-based policies to novel task variations.
We propose a hierarchical approach, where an upper-level policy learns to adapt the BT-based lower-level control policy, using trial-and-error experience collected on a small set of sample task variations.
The key concept is conditioning the upper-level policy on a \textit{context} vector that describes the possible task variations, enabling the policy to adjust the robot's behavior accordingly.
We train the upper-level policy in a sample-efficient manner, by adapting online Reinforcement Learning (RL) to operate within the BT structure.
This enables the policy to exploit prior experience when approaching new task variations during training.
We evaluate the efficiency of our approach in simulation, showing that the convergence time is independent of the number of task variations considered during training.
This enables training on large sets, improving generalization to task variations different from the ones used during training.
Moreover, we demonstrate the efficiency and effectiveness of our approach by training a policy directly on a Franka Emika Panda robot.


\section{Related Work}
Recent works have explored different BT-based policies, combining BTs with diverse low-level controllers to achieve complex multi-step tasks.
Rovida et al.~\cite{8594319} combine BTs with Motion Generators to model skills for contact-rich tasks, such as inserting a peg into a hole.
In~\cite{9907864}, we exploit a Stack-of-Tasks control strategy to decompose a robot skill (e.g., picking an object) into a set of prioritised tasks, explicitly and concurrently handled by different BT nodes.

Manually configuring the nodes of a BT-based policy for a task can be challenging.
Most methods introduce parameters in the leaf nodes of a predefined BT structure, such as intermediate goal poses of a robot's end-effector in a manipulation scenario.
These parameters can then be optimized using Policy Search methods~\cite{ps_book,ps_fast_survey} to achieve the desired robot behavior.
Mayr et al.~\cite{learn_params_bts} use Covariance Matrix Adaption Strategy to optimize the parameters of the BT-based policy proposed in~\cite{8594319}, relying on a manually designed BT structure.
They train in simulation, exploiting domain randomization techniques~\cite{domain_randomization} to bridge the sim-to-real gap.
Styrud et al.~\cite{styrud2024bebop} generate BTs by building a reactive tree structure with a planner, and then optimizing the parameters with Bayesian Optimization.
The above methods achieve good results under controlled and static conditions.
Building upon~\cite{learn_params_bts}, Ahmad et al.~\cite{perf} tackle adapting BT-based policies to task variations.
They train a Gaussian process support vector machine combined model on a dataset of task variations paired with optimal parameters.
During inference, the learned model predicts optimal parameters for unseen variations on the fly.
A key limitation is the training dataset construction, as it requires an optimization problem to be solved from scratch for each task variation (for example using~\cite{learn_params_bts,styrud2024bebop}).
In contrast, our approach avoids this cost by using online RL and sharing experience between task variants during training.

Our work can be viewed through the lens of contextual RL~\cite{contextual_rl,contextual_rl2}, which assumes underlying environment variations can be captured by a \textit{context}, informing a generalizable agent to adapt its behavior accordingly.
A naive approach concatenates the context directly with the observation~\cite{contextual_rl4,contextual_rl5,contextual_rl6,contextual_rl7}.
More advanced techniques range from learning separate representations for context and state before concatenation~\cite{contextual_rl8} to using a hypernetwork to adapt the primary network weights to the context~\cite{beukman_context}.
While these methods assume agents can observe the context, other works have focused on inferring it instead~\cite{contextual_rl4,contextual_rl9}.
In our work, we assume the context is observable and concatenate it with the observation.


\section{Learning of adaptive parameterized Behavior Tree-based policies}
In this paper, we assume that we are given a predetermined BT policy, composed of a number of internal nodes, which dictate execution flow, and leaf nodes, which include Action Nodes (ANs) that interact with a low-level robot controller and Condition Nodes that check propositions.
Further, some of the Action Nodes in the tree are parametrized, allowing us to set different values and locally adapt the robot behavior.
\begin{figure}[t!]
\centering
\includegraphics[width=0.97\linewidth]{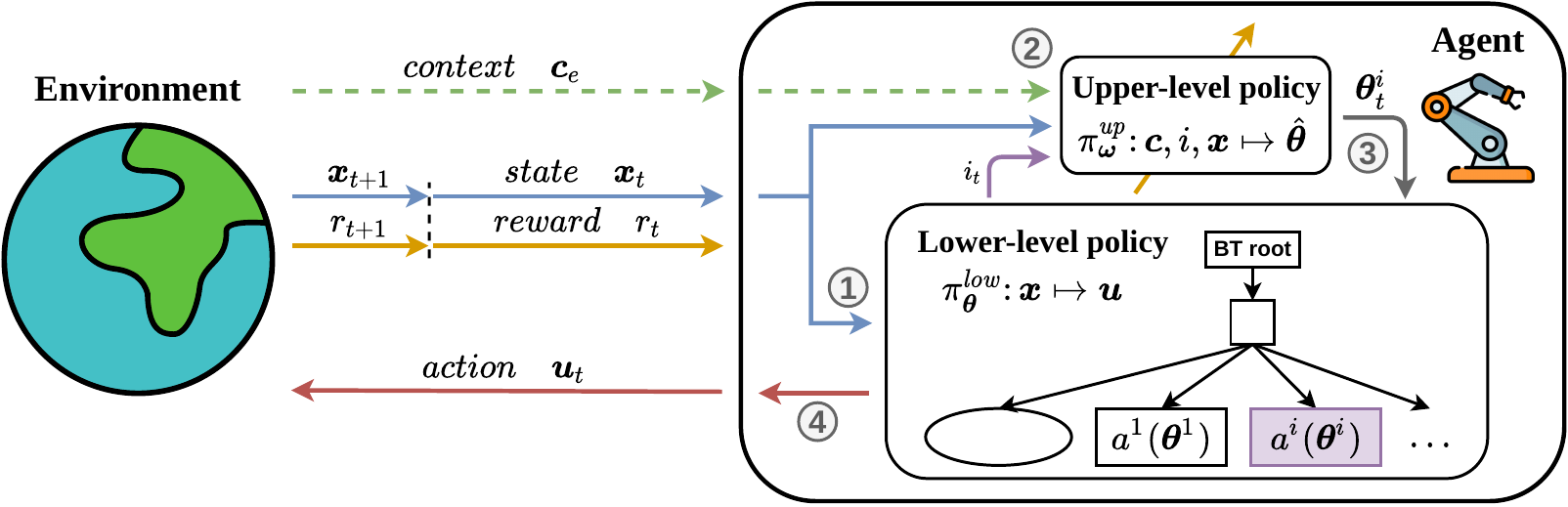}
\caption{Overview of our method for learning an agent that adapts to task variations. We propose a hierarchical approach, where an upper-level policy $\pi^{\mathit{up}}_{\boldsymbol{\omega}}$ selects a set of parameters that are used by a lower-level BT-based policy $\pi^{\mathit{low}}_{\boldsymbol{\theta}}$ to control the robot. The upper-level policy is conditioned on a context vector $\boldsymbol{c}$ encoding task variations, to adjust the robot's behavior accordingly. The environment provides an episodic context $\boldsymbol{c}_e$ and the state $\boldsymbol{x}_t$. $\pi^{\mathit{low}}_{\boldsymbol{\theta}}$ selects an Action Node $a^i$ to execute ($\textbf{1}$), and queries $\pi^{\mathit{up}}_{\boldsymbol{\omega}}$ for parameters $\boldsymbol{\theta}^i$ ($\textbf{2}$ and $\textbf{3}$).
Then, $a^i$ with parameters $\boldsymbol{\theta}^i$ is executed until completion or until the BT halts it ($\textbf{4}$).
The environment provides a reward $\boldsymbol{r}_t$, which is used to update the parameters $\boldsymbol{\omega}$ of the upper-level policy and the process is repeated.
}%
\label{fig:approach}
\end{figure}
Following terminology in~\cite{ps_book}, we adopt a hierarchical approach, where an upper-level policy selects a set of parameters that are used by a lower-level BT-based policy to control the robot (\Cref{fig:approach}).
The key idea is to condition the upper-level policy on a \textit{context} vector encoding task variations,
allowing it to select suitable parameters for different variations and adjust the robot's behavior accordingly. 
Using online RL, we learn the upper-level policy efficiently, exploiting prior experience on other task variations during training.

\subsection{Parameterized BT-based Control Policy}
\label{sec:lower-level-policy}
The lower-level policy is a BT-based control policy combining a controller with a BT.
The former is a closed-loop controller that directly modifies the robot state (e.g., in a manipulation task a Cartesian or impedance controller).
The latter is any BT whose ANs set commands to the controller to achieve a multi-step task.

Let $\pi^{\mathit{contr}}$ and $\pi^{\mathit{bt}}$ be the predefined controller and BT policies, respectively.
Let $\mathcal{A} = \{a^1, ..., a^n\}$ be the set of ANs in $\pi^{\mathit{bt}}$.
Considering sequential BT execution, $\pi^{\mathit{bt}}$ maps a state $\boldsymbol{x}$ to an AN $a \in \mathcal{A}$:
\begin{equation}
    \pi^{\mathit{bt}} \colon \boldsymbol{x} \mapsto a,
\end{equation}
where $\boldsymbol{x}$ can be composed of both internal states (e.g., joint velocities, end-effector pose) and external states (e.g., object locations).
The AN $a$ is then mapped by $\pi^{\mathit{contr}}$ to a robot control command $\boldsymbol{u}$:
\begin{equation}
    \pi^{\mathit{contr}} \colon a \mapsto \boldsymbol{u}.
\end{equation}
$\pi^{\mathit{low}}$ denotes the lower-level BT-based policy that combines the controller $\pi^{\mathit{contr}}$ and the BT $\pi^{\mathit{bt}}$ to obtain a control command $\boldsymbol{u}$, given a state $\boldsymbol{x}$:
\begin{equation}
    \pi^{\mathit{low}} \colon \boldsymbol{x} \mapsto \boldsymbol{u}.
\end{equation}
To adapt the robot behavior to task variations, we parameterize $\pi^{\mathit{bt}}$ with parameter vector $\boldsymbol{\theta}$, denoting it as $\pi^{\mathit{bt}}_{\boldsymbol{\theta}}$.
This allows us to define an upper-level policy that, operating in the parameter space of $\pi^{\mathit{bt}}_{\boldsymbol{\theta}}$, can adjust $\boldsymbol{\theta}$ to achieve the desired robot behavior.
In particular, we introduce parameters only in the ANs of the BT, assuming that the Condition Nodes are predefined.
Having non-parameterized Condition Nodes entails that, for a given input state $\boldsymbol{x}$, $\pi^{\mathit{bt}}_{\boldsymbol{\theta}}$ always selects the same AN $a$.
Conversely, a parameterized AN allows for modifications to the parameters to directly affect the commands sent by the BT to the controller.
As an example, in a robot manipulation scenario, these parameters may represent the target position and orientation of the end-effector. 
Formally, let $\mathcal{A} = \{a^1(\boldsymbol{\theta}^1), \ldots, a^n(\boldsymbol{\theta}^n)\}$ be the set of parameterized ANs in $\pi^{bt}_{\boldsymbol{\theta}}$, such that:
\begin{itemize}
    \item $\boldsymbol{\theta}^i \in \mathbb{R}^{m^i} \cup \{\emptyset\}$ is the parameter vector of AN $a^i$, with $i = 1, \ldots, n$;
    \item $\boldsymbol{\theta} = [ \boldsymbol{\theta}^1, \ldots, \boldsymbol{\theta}^n ] \in \mathbb{R}^m$ is the stacked parameter vector;
    \item and $\sum_{i=1}^{n} m^i = m$ is the dimensionality of $\boldsymbol{\theta}$.
\end{itemize} 
\noindent $\pi^{\mathit{low}}_{\boldsymbol{\theta}}$ denotes the resulting parameterized lower-level BT-based policy, where $\boldsymbol{\theta}$ is the parameter vector of $\pi^{bt}_{\boldsymbol{\theta}}$.

\subsection{Upper-level Policy}
\label{sec:upper-level-policy}
The upper-level policy selects parameters $\boldsymbol{\theta}$, used by the lower-level BT-based policy to control the robot and execute the task.
The two main design concepts in our upper-level policy are: \textit{dynamic} and \textit{step-based} parameter selection.

\textbf{Dynamic parameter selection}.
A fixed parameter choice would constrain the robot to a single, invariant behavior, which may be suboptimal or fail for even slight task variations~\cite{beukman_context}.
Instead,  we adjust the robot's behaviour in response to task variations, by making the upper-level policy dynamically select parameters for the lower-level policy.
To achieve this, we condition the upper-level policy on a \textit{context} vector $\boldsymbol{c} \in \mathbb{R}^c$ encoding the \textit{episodic} task variations~\cite{ps_book}, i.e., all variables that remain unchanged within a single task instance but may vary across multiple instances.
For example, in an obstacle avoidance task, this could represent the obstacle size and position.
Being \textit{context}-conditioned, the upper-level policy can distinguish between task variations and select parameters specifically.

\textbf{Step-based parameter selection}.
Given the episodic context, one approach to achieving task variation adaptation would be to select all AN parameters $\boldsymbol{\theta}$ at the beginning of each episode.
However, this choice is suboptimal for two reasons.
Firstly, the action space dimensionality of the upper-level policy scales linearly with the number of ANs.
Secondly, during learning, the upper-level policy receives only episodic rewards, reflecting overall performance across the episode.
These motivations lead us to define an upper-level policy that selects parameters only for the currently running AN.
Since each AN parameter vector may have a different number of components $m^i$, we design the upper-level policy to return a vector $\hat{\boldsymbol{\theta}} \in \mathbb{R}^p$, where $p$ is the maximum number of components among all the parameter vectors associated with the ANs:
\begin{equation}
    p = \max\{m^1, \ldots, m^n\} < \sum_{i=1}^{n} m^i = m.
\end{equation}
We condition the upper-level policy on the index $i$ of the AN $a^i$ selected by $\pi^{\mathit{low}}_{\boldsymbol{\theta}}$, allowing the policy to select parameters $\hat{\boldsymbol{\theta}}$ suitable for $a^i$.
We include $\boldsymbol{x}$ in the state, to differentiate between states with the same context $\boldsymbol{c}$ and index $i$.

The resulting upper-level policy $\pi^{up}_{\boldsymbol{\omega}}$ is defined as follows:
\begin{equation}
    \label{eq:up_policy}
    \pi^{\mathit{up}}_{\boldsymbol{\omega}} \colon \boldsymbol{c}, i, \boldsymbol{x} \mapsto \hat{\boldsymbol{\theta}},
\end{equation}
where $\boldsymbol{\omega}$ is the policy parameter vector, $\boldsymbol{c} \in \mathbb{R}^c$ the \textit{context} vector, $i \in \mathbb{R}$ the AN index selected by $\pi^{\mathit{bt}}_{\boldsymbol{\theta}}$, $\boldsymbol{x}$ the environment state, and $\hat{\boldsymbol{\theta}} \in \mathbb{R}^p$ the policy action.

\begin{algorithm}[t!]
\label{alg:learning}
\small
\caption{Learning of adaptive parameterized Behavior Tree-based policies.}
\SetAlgoNoLine
Initialize parameter vector $\boldsymbol{\omega}$ and replay buffer $\mathcal{R} \gets \emptyset$. \\
\For{$e=1, \dots, episodes$}{
    Receive observation $\boldsymbol{x}_1$ and context $\boldsymbol{c}_e$ \label{line:context}\\
    Select Action Node $a_1= \pi^{bt}_{\boldsymbol{\theta}} (\boldsymbol{x}_1)$ \label{line:action-node-selection} \\
    \For{$t=1, \dots, steps$ and $!done$}{
        Select parameter vector $\hat{\boldsymbol{\theta}}_t = \pi^{up}_{\boldsymbol{\omega}}(\boldsymbol{c}_e, [a_t]_{id})$ \label{line:parameters-selection} \\
        Select commands $\boldsymbol{u}_t = \pi^{contr}(\boldsymbol{x}_t, a_t(\hat{\boldsymbol{\theta}}_t))$ \\
        Execute commands $\boldsymbol{u}_t$, observe $r_t$ and $\boldsymbol{x}_{t+1}$ \label{line:execution}\\
        \eIf{done}{
            $a_{t+1} \gets {none}$ \\
        }{
            Select next Action Node $a_{t+1}= \pi^{bt}_{\boldsymbol{\theta}} (\boldsymbol{x}_{t+1})$ \label{line:action-node-selection2}\\
        }
        $\mathcal{R} \gets \mathcal{R} \cup \{((\boldsymbol{c}_e, [a_t]_{id}), \hat{\boldsymbol{\theta}}_{t}, r_t, (\boldsymbol{c}_e, [a_{t+1}]_{id}))\}$ \label{line:replay-buffer}\\

        \For{$j=1, \dots, iterations$}{
            Sample a random minibatch from $\mathcal{R}$ \label{line:sample-transition} \\
            Update $\boldsymbol{\omega}$ \label{line:optimize-policy}
        }
    }
  
}
\end{algorithm}

\subsection{Learning}
\label{sec:learning}

We use online RL to efficiently learn the upper-level policy $\pi^{up}_{\boldsymbol{\omega}}$.
During learning the agent is presented with different task variants at every episode, chosen from a limited training set.
This enables fast convergence to a generalizable policy, leveraging prior experience when facing new variations during training.

Algorithm \ref{alg:learning} describes the overall learning procedure, applicable to any off-policy RL algorithm.
At the beginning of each episode, the environment provides the global context ${\boldsymbol{c}_e}$ (line \ref{line:context}).
At each time-step $t$, the BT policy $\pi^{\mathit{bt}}_{\boldsymbol{\theta}}$ selects the AN $a_t$ to perform, based on the current environment state $\boldsymbol{x}_t$ (lines \ref{line:action-node-selection} and \ref{line:action-node-selection2}).
The upper-level policy $\pi^{\mathit{up}}_{\boldsymbol{\omega}}$ selects optimal parameters $\hat{\boldsymbol{\theta}_t}$ given context $\boldsymbol{c}_e$ and the index of the selected AN $a_t$ (line \ref{line:parameters-selection}).
We use $[.]_{\mathit{id}}$ to denote the operator that maps from $a_t$ to index $i_t$.
Last, $\pi^{\mathit{contr}}$ executes $a_t$ with parameters $\hat{\boldsymbol{\theta}_t}$ until completion or until halted by the BT when a condition is no longer met (line \ref{line:execution}).
Optimization of $\pi^{\mathit{up}}_{\boldsymbol{\omega}}$ relies on a replay buffer $\mathcal{R}$, where we collect transitions as:
\begin{equation}
    \label{eq:transition}
	((\boldsymbol{c}_e, [a_t]_{\mathit{id}}), \hat{\boldsymbol{\theta}}_{t}, r_t, (\boldsymbol{c}_e, [a_{t+1}]_{\mathit{id}})),
\end{equation}
where $(\boldsymbol{c}_e, [a_t]_{\mathit{id}})$ and $(\boldsymbol{c}_e, [a_{t+1}]_{\mathit{id}})$ are the states for $\pi^{\mathit{up}}_{\boldsymbol{\omega}}$ at time-step $t$ and $t+1$ respectively, $\hat{\boldsymbol{\theta}}_{t}$ the action selected by $\pi^{\mathit{up}}_{\boldsymbol{\omega}}$ at time-step $t$, and $r_t$ the reward provided by the environment (line \ref{line:replay-buffer}).
The policy $\pi^{up}_{\boldsymbol{\omega}}$ is optimized by sampling a minibatch of transitions from $\mathcal{R}$ (lines \ref{line:sample-transition}-\ref{line:optimize-policy}).


\section{Evaluation}
In \Cref{sec:obstacle-avoidance}, we evaluate our approach in simulation, while in \Cref{sec:pivoting} we train a policy directly on a physical robot.
As BT implementation we use the BehaviorTree.CPP library \cite{bt}.
For learning, we employ the Soft Actor-Critic (SAC) algorithm~\cite{sac}, in its implementation provided by \textit{Stable-Baselines3} \cite{stable-baselines3}.

\subsection{Obstacle Avoidance}
\label{sec:obstacle-avoidance}

\begin{figure}[t!]
    \centering
    \subfigure[]{\includegraphics[width=0.35\textwidth]{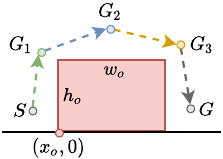}\label{fig:obstacle_avoidance_task}}
    \hspace{1cm}
    \subfigure[]{\includegraphics[width=0.41\textwidth]{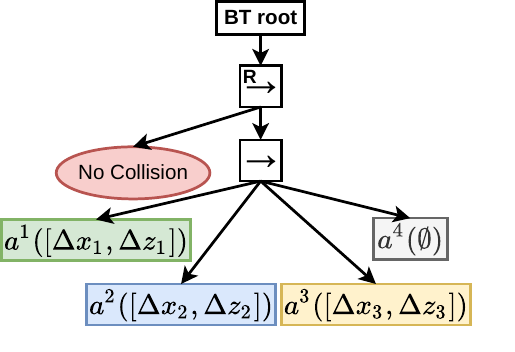}\label{fig:obstacle_avoidance_bt}} 
    \caption{Obstacle avoidance task. \ref{fig:obstacle_avoidance_task} The objective for a robotic arm is to move its end-effector between predetermined start and goal positions ($S$ and $G$ respectively) while avoiding a static obstacle. Task variations arise from different obstacle heights $h_o$, widths $w_o$, and positions in a horizontal direction $x_o$. \ref{fig:obstacle_avoidance_bt} We design a BT policy $\pi^{bt}_{\boldsymbol{\theta}}$ with $4$ Action Nodes, each performing a linear motion. $\Delta x_i$ and $\Delta z_i$ are the goal relative coordinates for each motion w.r.t the current position. The last Action Node is not parameterized, as the goal location $G$ is predetermined. \textit{\textbf{R}} denotes a \textit{reactive} control node~\cite{bts_book}, continuously checking for collisions during motion.}%
\end{figure}

\textbf{Task description.}
The objective for a robotic arm is to move its end-effector between predetermined start and goal positions while avoiding collisions with a static obstacle in the workspace.
Similarly to~\cite{perf}, we introduce task variations by alternating between different obstacle heights $h_o$, widths $w_o$, and positions in a horizontal direction $x_o$.
While the method in~\cite{perf} controls the end-effector of a 7-DoF robot, their task reduces to controlling a point robot on a 2D plane, due to the absence of other relevant obstacles or undesirable regions. 
For simplicity and to enable rapid training, we solve the underlying 2-DoF task version directly.
We implement a simple simulation environment featuring a 2D state space $[x_t,z_t] \in \mathbb{R}^2$ and a 2D action space $[\Delta x_t, \Delta z_t] \in \mathbb{R}^2$ that allows a low-level control policy to command linear motions between different points in the plane (\Cref{fig:obstacle_avoidance_task}).

\textbf{Lower-level policy.}
We design a BT policy $\pi^{\mathit{bt}}_{\boldsymbol{\theta}}$ with $4$ ANs (\Cref{fig:obstacle_avoidance_bt}), each performing a linear motion by interpolating between the current and the desired position.
The first three ANs are parameterized with the $x$ and $z$ coordinates of the commanded intermediate goals.
The last AN is not parameterized, as the goal location $G$ is predetermined.
It follows that $\boldsymbol{\theta}^i = [x_i, z_i] \in \mathbb{R}^2$, with $i = 1, \ldots, 3$, and $\boldsymbol{\theta} = [x_1, z_1, x_2, z_2, x_3, z_3] \in \mathbb{R}^6$.
This results in a $6D$ parameter space for $\boldsymbol{\theta}$, identical to the action space dimensionality in \cite{perf}.
The task is successfully completed when the final goal location $G$ is reached without collisions.

\textbf{Upper-level policy.}
We learn the following policy:
\begin{equation}
    \label{eq:upper-level-ostacle-avoidance}
	\pi^{\mathit{up}}_{\boldsymbol{\omega}} \colon [h_o, w_o, x_o], i, [x, z] \mapsto [\Delta x, \Delta z],
\end{equation}
where:
\begin{itemize}
    \item $[h_o, w_o, x_o]$ is the \textit{context} vector $\boldsymbol{c}$, with $h_o, w_o, x_o$ being the height, width and $x$ coordinate of the bottom left corner of the obstacle, respectively;
    \item $[x, z]$ is the $2D$ robot position;
    \item $[\Delta x, \Delta z]$ is the output vector $\hat{\boldsymbol{\theta}}$ of the policy, with $\Delta x$ and $\Delta z$ being the goal relative coordinates for each motion w.r.t the end-effector position.
\end{itemize}

\textbf{Reward function.} We employ the same reward function as in \cite{learn_params_bts, perf}, which is a weighted sum of three components: \textit{goal distance}, \textit{collision avoidance}, and \textit{task completion}.
The \textit{goal distance} component $r_g$ rewards proximity to the goal position, while the \textit{collision avoidance} component $r_c$ penalizes proximity to the obstacle.
For more details on these two components, refer to \cite{learn_params_bts}.
Lastly, the \textit{task completion} component rewards successful task completion by optimizing the number of steps taken.
It is defined as \(r_s (\mathit{n\_steps}) = \mathit{max\_steps} - \mathit{n\_steps}\), where \(\mathit{max\_steps} = 200\) and \(\mathit{n\_steps}\) is the total number of way-points interpolated from \(S\).
We compute $r_g$ and $r_c$ for each interpolated way-point until reaching the goal or colliding with an obstacle, while $r_s$ is considered only upon reaching the goal position.
This ensures the policy focuses first on finding a collision-free path to the goal and then optimizing the path length.

\begin{figure}[t!]
    \centering
    \subfigure[]{\includegraphics[width=0.4\textwidth]{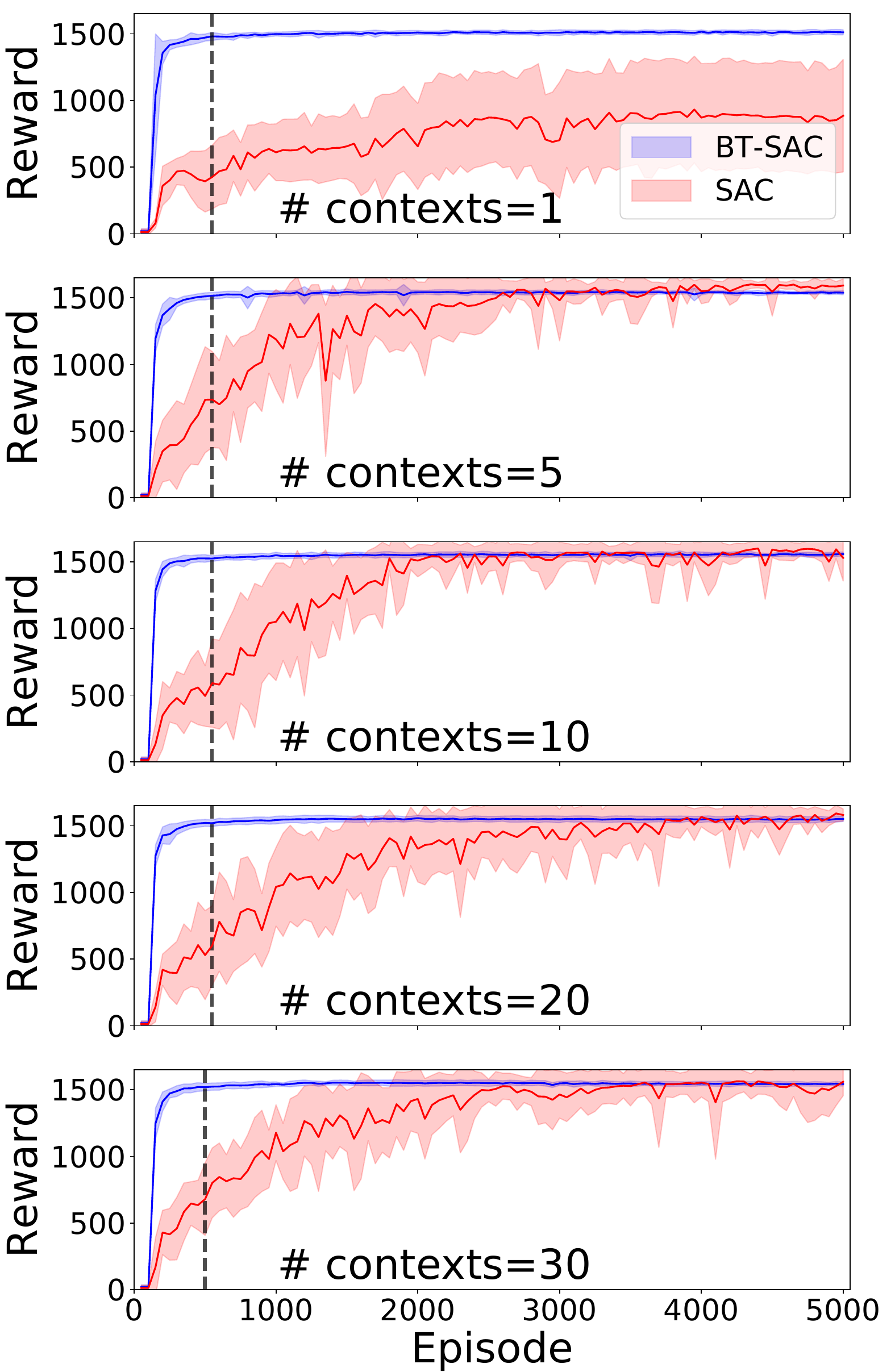}\label{fig:obstacle-avoidance-curves}}
    \subfigure[]{\includegraphics[width=0.4\textwidth]{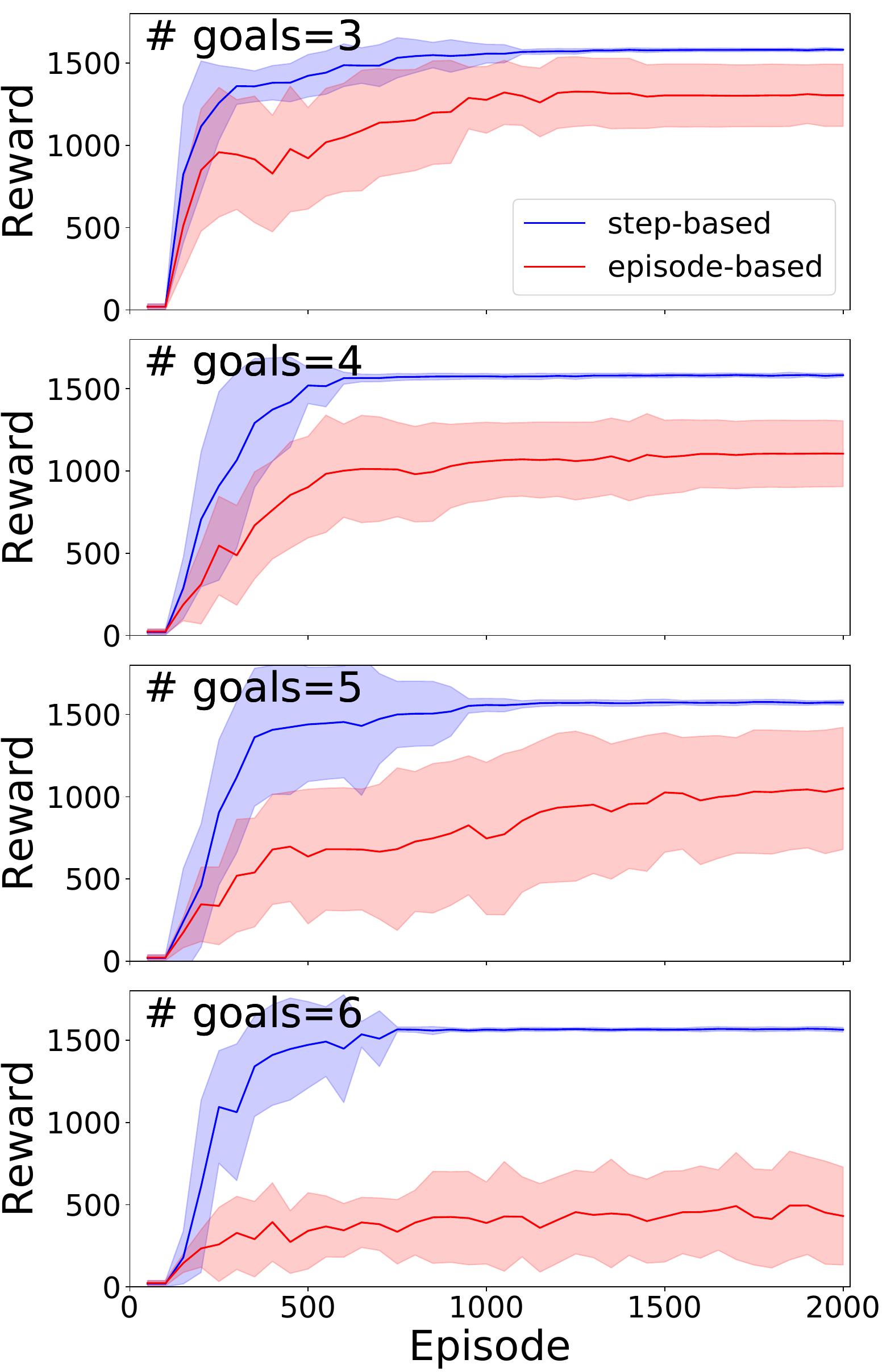}\label{fig:obstacle-avoidance-curves-episode}}
    \caption{Learning curves for the obstacle avoidance task obtained by periodically evaluating policies on all training contexts. The solid line and shaded region represent the mean and standard deviation, respectively ($10$ replicates). \Cref{fig:obstacle-avoidance-curves} compares our BT-based policy with a standard SAC policy on an increasing number of training contexts. The dashed line indicates our policy convergence (i.e., reward improvement over the last $150$ episodes < $2\%$). \Cref{fig:obstacle-avoidance-curves-episode} compares our step-based upper-level policy with an episode-based one on an increasing number of intermediate goals (ANs in \Cref{fig:obstacle_avoidance_bt}).}%
    \vspace{-0.3cm}
\end{figure}

\textbf{Training.}
To assess the scalability of our approach, we train $5$ different policies, increasing the number of task variations considered during training (\Cref{fig:obstacle-avoidance-curves}).
We use the same task variation ranges considered in \cite{perf}, employing Latin hypercube sampling.
As a baseline, we train a \textit{SAC} policy that directly selects local actions without utilizing any prior task knowledge encoded in the BT.
This baseline allows us to highlight the efficiency advantage of our BT-based approach (\textit{BT-SAC}) and to establish a benchmark for evaluating the generalization capability of our method in terms of reward-based performance.

Furthermore, to investigate the efficiency advantage of a step-based parameter selection, we compare our upper-level policy with an episode-based one (\Cref{fig:obstacle-avoidance-curves-episode}).
This policy maps the context $[h_o, w_o, x_o]$ to the parameters $\boldsymbol{\theta}$ of all ANs at the beginning of each episode, receiving a cumulative episodic reward for the entire executed trajectory.
To emphasize the difference between the policies, we increase the task difficulty by introducing a forbidden area $10\mathit{cm}$ above the obstacle, triggering BT failure upon robot entry.
We compare the policies on an increasing number of intermediate goals (i.e., ANs in \Cref{fig:obstacle_avoidance_bt}), to evaluate their scalability on the number of components of $\boldsymbol{\theta}$.
Specifically, we consider $3$, $4$, $5$, and $6$ intermediate goals, resulting in a $6D$, $8D$, $10D$, and $12D$ parameter space for $\boldsymbol{\theta}$, respectively, while keeping the number of training contexts fixed at $20$.

\textbf{Evaluation.}
We assess to what extent the policies learned on a different number of contexts in \Cref{fig:obstacle-avoidance-curves} can adapt to task variations not encountered during training.
For this purpose, we evaluate their performance on $20$ novel task variations, sampled using Latin Hypercube Sampling from the same training ranges with a different seed.
For both our \textit{BT-SAC} and \textit{SAC}, we evaluate the final models obtained at the end of training.
Additionally, as baselines, we train three new policies on the $20$ validation task variations: \textit{BT-SAC}, \textit{SAC} and a BT-based policy optimized through Bayesian Optimization (\textit{BT-BO}) as in~\cite{styrud2024bebop}, in its implementation provided by~\cite{bo}.
These baselines show the performance achievable if the policies were trained specifically on the validation set.
We are unable to conduct a fair direct comparison of reward-based performance with~\cite{perf}, due to the limited details provided on their training implementation.
However, \textit{BT-BO} enables an indirect comparison with~\cite{perf}, as their training data is generated via black-box optimization, and their model can attain, at most, the reward achieved by the optimizer.
The evaluation results are shown in \Cref{tab:obstacle-avoidance}.

\textbf{Analysis.}
Analysis of the learning curves reveals two main key findings.
Firstly, \Cref{fig:obstacle-avoidance-curves} shows that our BT-based approach exhibits consistent convergence behavior, reaching convergence within a similar number of episodes irrespective of the considered number of task variations.
This provides a clear advantage compared to the state-of-the-art~\cite{perf}, which scales linearly with the number of task variations considered during training, as it requires solving an optimization problem from scratch for each training task variation.
In contrast, our method enables the policy to exploit prior experience when facing new task variations during training, facilitating training on larger sets to improve its generalization ability.
We also observe that our approach converges faster and with less variance than \textit{SAC}, due to the prior knowledge encoded in the BT --- namely, the task decomposition into three linear motions --- which simplifies learning.
Secondly, \Cref{fig:obstacle-avoidance-curves-episode} shows the efficiency advantage of a step-based parameter selection.
The step-based policy exhibits faster convergence and lower variance, while the episode-based policy suffers from the curse of dimensionality, ultimately failing to learn to complete the task with $6$ intermediate goals.

Inspection of \Cref{tab:obstacle-avoidance} indicates a clear trend: as the number of task variations increases, so do the generalization capabilities of the policies.
In contrast to \textit{SAC}, our \textit{BT-SAC} policies trained on $20$ and $30$ task variations achieve performance comparable to the corresponding ones trained directly on the validation set.
Both policies trained on $30$ contexts do not appear to yield additional advantages, indicating an optimal basis of task variations beyond which further experience is redundant.
While simplifying learning, the BT prior knowledge limits the policy’s expressiveness, leading to suboptimal performance, justifying the higher reward observed with \textit{SAC}.
This observation is further supported by the \textit{BT-BO} performance, which shares the same underlying structure as \textit{BT-SAC} and also underperforms compared to \textit{SAC}.
Finally, although \textit{BT-BO} achieves marginally better performance than \textit{BT-SAC}, it shares a key limitation with the episode-based upper-level policy analyzed in \Cref{fig:obstacle-avoidance-curves-episode}: susceptibility to the curse of dimensionality.
This arises because it operates on the entire BT parameter space, leading to a decrease in performance with an increasing number of BT parameters.
This is also a limitation of~\cite{perf}, as their training data is likewise generated through black-box optimization.

\begin{table}[t!]
    \centering
    \scriptsize
    \renewcommand\arraystretch{1.2}
    \caption{Performance obstacle avoidance task. \textit{BT-SAC} refers to our approach.}
    \begin{tabular}{c|ccc|ccc}
        \multicolumn{1}{c}{\textbf{\# contexts}} & \multicolumn{3}{c}{\textbf{Reward}} & \multicolumn{3}{c}{\textbf{\# collisions}} \\ [0.5ex]
        & BT-SAC & SAC & BT-BO & BT-SAC & SAC & BT-BO \\
        \toprule
        1       & 1163 $\pm$ 65 & 653 $\pm$ 185 & -     & 5.4 $\pm$ 1.0 & 7.2 $\pm$ 6.3    & -     \\ 
        5       & 1417 $\pm$ 122 & 1370 $\pm$ 251 & -    & 2.2 $\pm$ 1.8 & 3.6 $\pm$ 3.5   & -   \\
        10      & 1443 $\pm$ 142 & 1550 $\pm$ 130 & -     & 1.8 $\pm$ 2.2 & 1.0 $\pm$ 1.7  & -    \\
        20      & 1560 $\pm$ 21 & 1590 $\pm$ 73 & -     & 0.0 $\pm$ 0.0 & 0.4 $\pm$ 0.8    & -   \\
        30      & 1567 $\pm$ 16 & 1577 $\pm$ 98 & -     & 0.0 $\pm$ 0.0 & 0.0 $\pm$ 0.0    & -  \\
        \midrule
        \textit{Trained on validation set}    & \textit{1559 $\pm$ 13}     & \textit{1632 $\pm$ 31}    & \textit{1608 $\pm$ 4} & \textit{0.0 $\pm$ 0.0} &\textit{ 0.0 $\pm$ 0.0} & \textit{0.0 $\pm$ 0.0}      \\
        \bottomrule
    \end{tabular}
    \label{tab:obstacle-avoidance}
    \vspace{-0.2cm}
\end{table}

\subsection{Pivoting}
\label{sec:pivoting}

\begin{figure}[t!]
    \centering
    {\includegraphics[width=0.65\linewidth,trim={0 0 0 2.5cm},clip]{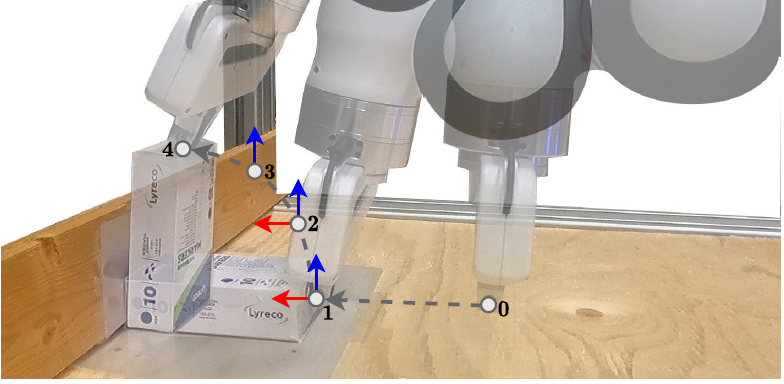}} 
    \caption{Illustration of an object pivoting task being executed on a Franka Emika Panda 7-DoF manipulator. We perform 4 motions in the $x$-$z$ plane, by commanding goal relative coordinates w.r.t. the current end-effector position. Solid arrows denote directions along which the goal relative coordinate is learned. The grey dashed line shows the overall trajectory.}
    \label{fig:pivoting}
\end{figure}

\textbf{Task description.}
The objective for a robotic arm is to manipulate an object adjacent to a wall, using the wall as a pivot to rotate the object by $90$ degrees (\Cref{fig:pivoting}).
Recent works highlight the challenge of learning policies that generalize effectively across diverse variations of this task \cite{yang2023learning, zhou2022learning, 6907062, 10.1007/978-3-030-28619-4_39}.
We collect multiple boxes, constructing a dataset that comprises $16$ different box sizes, whose range spans from $12.2\mathit{cm}$ to $20.5\mathit{cm}$, with weights between $15\mathit{g}$ and $87\mathit{g}$.

\textbf{Lower-level policy.}
As low-level controller $\pi^{\mathit{contr}}$, we use a Cartesian impedance controller, commonly used for contact-rich tasks as it regulates interaction forces and motions, ensuring stable and compliant interaction with the environment~\cite{impedance-contr}.
We design a BT $\pi^{\mathit{bt}}_{\boldsymbol{\theta}}$ to command $4$ linear motions, each in a different AN (\Cref{fig:pivoting}).
The first AN is not parameterized and aims to detect the box size.
Starting from a predetermined position, the end-effector is commanded to advance linearly along the $x$-axis until encountering the object.
To accomplish this detection, we employ a Condition Node that evaluates whether the force exerted on the end-effector exceeds a specified threshold.
Once contact with the object is detected, we use the $x$ coordinate of the end-effector position as a context, allowing the upper-level policy to distinguish between different box sizes.
The following $2$ ANs are parameterized with the $x$ and $z$ coordinates of the intermediate goals to be commanded, while the last AN is parameterized only with the $z$ coordinate.
This is because the final goal coordinate $x_4$ does not need to be learned, as it does not depend on the box size and can therefore be fixed.
It follows that $\boldsymbol{\theta}^1 \in \{\emptyset\}$, $\boldsymbol{\theta}^i = [x_i, z_i] \in \mathbb{R}^2$, with $i = 2, 3$, and $\boldsymbol{\theta}^4 = [z_4] \in \mathbb{R}$, thus $\boldsymbol{\theta} = [x_2, z_2, x_3, z_3, z_4] \in \mathbb{R}^5$.
The task is successfully completed when the end-effector reaches $x_4$ without having dropped the box.
For this purpose, we design a Condition Node, that triggers the BT failure if the box is dropped.

\textbf{Upper-level policy.}
We learn the following policy:
\begin{equation}
	\pi^{up}_{\boldsymbol{\omega}} \colon [x_1], i, [x, z] \mapsto [\Delta x, \Delta z],
\end{equation}
where:
\begin{itemize}
    \item $[x_1]$ is the \textit{context} vector, with $x_1$ being the $x$ coordinate of the end-effector when the box is detected;
    \item $[x, z]$ is the $2D$ end-effector position;
    \item $[\Delta x, \Delta z]$ is the policy output vector $\hat{\boldsymbol{\theta}}$, with $\Delta x$ and $\Delta z$ being the goal relative coordinates for each motion w.r.t the initial end-effector position.
\end{itemize}

\textbf{Reward function.}
We use a sum of three components: \textit{goal distance}, \textit{exploration incentive}, and \textit{task completion}.
The \textit{goal distance} component rewards proximity to the goal position, as in~\cite{learn_params_bts}, but computes the distance only between the last motion point and the goal coordinate $x_4$, independent of the box size.
To encourage exploration, the \textit{exploration incentive} component penalizes trajectories shorter than $1\mathit{cm}$ by adding a fixed negative reward.
Not being interested in optimizing the trajectory length, we consider a fixed positive reward upon task completion, instead of the path length-dependent one used in \Cref{sec:obstacle-avoidance}.

\textbf{Training.}
We train $3$ policies on 4 different box sizes each to explore the generalization capabilities of our approach.
We assess interpolation by training a policy $\pi_{\mathit{i}}$ on a set spanning the entire context range with the smallest, largest, and two intermediate sizes.
To evaluate extrapolation, we train the other two policies on extreme context ranges, using the $4$ smallest sizes for $\pi_{\mathit{es}}$ and the $4$ largest for $\pi_{\mathit{el}}$.
We train each policy from scratch on a Franka Emika Panda 7-DoF manipulator by cycling through its training set and positioning a different box in the same location at each episode.
For each policy, we perform $80$ random exploration steps plus $80$ training steps, totalling about an hour of training.

\begin{table}[t!]
    \centering
    \footnotesize
    \caption{Pivoting task. Highlighted cells indicate training boxes.}
    \renewcommand\arraystretch{1.2}
    \hspace{-0.4cm}
    \begin{tabular}{>{\centering}m{0.05\textwidth}>{\centering}m{1.5em}>{\centering}m{1.5em}>{\centering}m{1.5em}>{\centering}m{1.5em}>{\centering}m{1.5em}>{\centering}m{1.5em}>{\centering}m{1.5em}>{\centering}m{1.5em}>{\centering}m{1.5em}>{\centering}m{1.5em}>{\centering}m{1.5em}>{\centering}m{1.5em}>{\centering}m{1.5em}>{\centering}m{1.5em}>{\centering}m{1.5em}>{\centering}m{1.5em}>{\centering\arraybackslash}m{3em}}
    
     & \rotatebox{55}{12.2} & \rotatebox{55}{12.5} & \rotatebox{55}{12.8} & \rotatebox{55}{13.0} & \rotatebox{55}{14.0} & \rotatebox{55}{14.5} & \rotatebox{55}{14.7} & \rotatebox{55}{15.2} & \rotatebox{55}{15.5} & \rotatebox{55}{15.7} & \rotatebox{55}{16.5} & \rotatebox{55}{17.0} & \rotatebox{55}{17.8} & \rotatebox{55}{18.6} & \rotatebox{55}{19.5} & \rotatebox{55}{20.5} &  \\ 
     \toprule

    $\pi_{\mathit{i}}$ & \cellcolor{green!25}\cmark & \cmark & \cmark & \cmark & \cmark & \cmark & \cellcolor{green!25}\cmark & \cmark & \cmark & \cmark & \cmark & \cellcolor{green!25}\cmark & \cmark & \cmark & \cmark & \cellcolor{green!25}\cmark & {\textbf{16/16}} \\

    \midrule
    
    $\pi_{\mathit{es}}$ & \cellcolor{green!25}\cmark & \cellcolor{green!25}\cmark & \cellcolor{green!25}\cmark & \cellcolor{green!25}\cmark & \cmark & \cmark & \cmark & \cmark & \cmark & \cmark & \cmark & \xmark & \xmark & \xmark & \xmark & \xmark & {\textbf{11/16}} \\
    
     $\pi_{\mathit{el}}$ & \xmark & \xmark & \cmark & \xmark & \cmark & \cmark & \cmark & \cmark & \cmark & \cmark & \cmark & \cmark & \cellcolor{green!25}\cmark & \cellcolor{green!25}\cmark & \cellcolor{green!25}\cmark & \cellcolor{green!25}\cmark & {\textbf{13/16}} \\

    \end{tabular}
    \label{tab:pivoting}
\end{table}

\textbf{Evaluation and analysis.}
We evaluate each policy on the entire dataset, with results in \Cref{tab:pivoting}.
All policies have learned to successfully flip the boxes provided during training (highlighted in green in Table~\ref{tab:pivoting}).
As expected, $\pi_{\mathit{i}}$ achieves the best generalization, completing the task for all the boxes in the dataset.
The other two policies show similar effectiveness in generalizing to distances of around $3/3.5\mathit{cm}$ from the training boxes.
$\pi_{\mathit{el}}$ performs slightly better in terms of success rate, due to the uneven distribution of box sizes in the dataset.

\section{Conclusion}
In this work, we introduce an efficient method for learning how to adapt BT-based control policies to unseen task variations.
We propose a hierarchical approach, where an upper-level policy selects a set of parameters, which are then used by the corresponding lower-level BT-based policy to control the robot.
By conditioning the upper-level policy on a context vector that encodes possible task variations, the upper-level policy can adjust the robot's behavior accordingly.
We exploit online RL to train the upper-level policy in a sample-efficient manner.
Experiments in simulation and on a physical robot demonstrate the efficiency and effectiveness of our approach.
One limitation is the assumption of non-parameterized Condition Nodes, which may hinder adaptation in some real-world scenarios.
Future work should explore integrating learnability into Condition Nodes for more adaptive robot behavior.

\begin{credits}
\subsubsection{\ackname} This work was supported in part by Industrial Graduate School Collaborative AI \& Robotics (CoAIRob), in part by the Swedish Knowledge Foundation under Grant Dnr:20190128, and the Knut and Alice Wallenberg Foundation through Wallenberg AI, Autonomous Systems and Software Program (WASP).
\subsubsection{\discintname}
The authors have no competing interests to declare that are relevant to the content of this article.
\end{credits}
%
%
%
%
\bibliographystyle{splncs04}
\bibliography{references}

\end{document}